\documentclass{article}

\usepackage{preprint}

\usepackage[utf8]{inputenc} 
\usepackage[T1]{fontenc}    
\usepackage{hyperref}       
\usepackage{url}            
\usepackage{booktabs}       
\usepackage{amsfonts}       
\usepackage{nicefrac}       
\usepackage{microtype}      
\usepackage{lipsum}
\usepackage{wrapfig}

\usepackage{booktabs} 
\usepackage{subfig}
\usepackage{amsmath,amssymb}
\usepackage{mathtools}
\usepackage[toc,page]{appendix}
\usepackage{enumitem}
\usepackage{graphics}
\usepackage{graphicx}
\usepackage{algorithm}
\usepackage{algorithmicx}
\usepackage[noend]{algpseudocode}
\algnewcommand{\LineComment}[1]{\State \(//\) #1}

\usepackage{adjustbox}
\usepackage{array}

\DeclareMathOperator{\EX}{\mathbb{E}}

\usepackage[font=small]{caption}

\title{Spiking Neural Predictive Coding for Continually Learning from Data Streams}

\author{
  Alexander Ororbia \\
  Department of Computer Science \\
  Rochester Institute of Technology \\
  Rochester, NY 14623 \\
  \texttt{ago@cs.rit.edu} \\
}

\begin{document}
\maketitle

\begin{abstract}
For energy-efficient computation in specialized neuromorphic hardware, we present spiking neural coding, an instantiation of a family of artificial neural models grounded in the theory of predictive coding. This model, the first of its kind, works by operating in a never-ending process of ``guess-and-check'', where neurons predict the activity values of one another and then adjust their own activities to make better future predictions. The interactive, iterative nature of our system fits well into the continuous time formulation of sensory stream prediction and, as we show, the model's structure yields a local synaptic update rule, which can be used to complement or as an alternative to online spike-timing dependent plasticity. In this article, we experiment with an instantiation of our model consisting of leaky integrate-and-fire units. However, the framework within which our system is situated can naturally incorporate more complex neurons such as the Hodgkin-Huxley model. Our experimental results in pattern recognition demonstrate the potential of the model when binary spike trains are the primary paradigm for inter-neuron communication. Notably, spiking neural coding is competitive in terms of classification performance and experiences less forgetting when learning from task sequence, offering a more computationally economical, biologically-plausible alternative to popular artificial neural networks. 
\end{abstract}



\section{Introduction}
\label{sec:intro}
Through the extraction of rich, expressive hierarchies of feature detectors from large samples of patterns, artificial neural networks (ANNs) have become popular, powerful models of data. ANNs continue to demonstrate their powerful function approximation abilities in an ever-growing variety of applications \cite{jo2015improving,park2015deep,rajkomar2018scalable,poplin2018prediction}, including many in computer vision \cite{krizhevsky2012imagenet,he2016identity},
speech \cite{hinton2012deep,oord2016wavenet}, and text processing \cite{ororbia2017deltarnn,devlin2018bert,ororbia2019likebaby}.
However, despite these recent successes, ANNs, in their current form, face some significant difficulties. These range from problems in effective implementation in (neuromorphic) hardware meant for real-time computation \cite{boahen2005neuromorphic,indiveri2011neuromorphics} to challenges in parameter optimization \cite{pascanu2013difficulty} to designing agents that generalize well across tasks and datasets \cite{thrun1995lifelong,zeng2019continual}.
The source of many of these issues might result from the fact that ANNs are only very loosely inspired by the brain. In essence, artificial neurons lack many of the actual mechanisms that underlie real biological neurons that, when taken en masse, give rise to the complex cognitive functioning and behavior that is characteristic of human agents.
If the hope is to one day craft neural systems that exhibit such behavior, a fruitful direction would be to bridge the gap between artificial and real neurons. 

Recently, a line of ANN research has begun to more deeply investigate this hypothesis, focusing on formalizing and implementing various types of neural computation and learning mechanisms currently known in cognitive neuroscience. This brand of connectionism, also known as biologically- or neurocognitively-plausible learning, has already begun to show that by modeling and integrating neurobiological mechanisms, e.g., lateral inhibition \cite{grossberg1987competitive,olshausen1997sparse,adesnik2010lateral,szlam2011structured,krotov2019unsupervised,ororbia2019lifelong}, neurotransmission \cite{meier1991neurotransmitters,doya2002metalearning,angela2005uncertainty}, local weight update rules \cite{jaderberg2017decoupled,ororbia2018conducting,ororbia2019biologically}, we might be either able to generalize differently and more robustly or, at the very least, side-step some of the issues related to classical ANN learning algorithms, i.e., back-propagation of errors (backprop) \cite{glorot2010understanding}.
This body of work continues to slowly generate evidence that, by more faithfully modeling how real neurons compute and conduct credit assignment, we might develop agents capable of processing complex patterns and operating in noisy environments.

One of the key differences between the neurons in ANNs and actual neurons is in how information is communicated across the individual units of the system. ANN neurons communicate with continuously graded values, typically requiring activation/transfer functions that are differentiable. In contrast, inter-neuron communication in real-world neurons is performed through the broadcasting of action potentials, creating series of spikes known as spike trains \cite{de2007correlation}. These spike trains are sparse in time, yet each spike carries a great deal of information. This serves as one primary inspiration for the spiking neural network (SNN) (a third generation neural network model), often used in computational neuroscience, which encodes sensory information content through the precise timing of spikes \cite{maass1997networks}.
The sparse, spike-based communication inherent to SNNs is, in fact, one of their primary strengths.
ANNs, on the other hand, require the use of energy-intensive, top-of-the-line graphics processing units (GPUs) for effective training while SNNs consume dramatically less energy through the use of high-information content spike trains, facilitating the construction of very fast, energy-efficient hardware to support their processing, i.e., neuromorphic hardware \cite{furber2014spinnaker,merolla2014truenorth,davies2018loihi}.
Importantly, the low energy-consumption, hardware-friendly nature of SNNs make them ideally suited for use in pattern processing and adaptation in complex real-time systems, such as robotic agents \cite{hagras2004evolving,rueckert2016recurrent} or self-driving cars \cite{hwu2017self}.

However, despite their potential, SNN research and simulation is still in its early stages (in contrast to second-generation ANNs) where the adjustment of the synaptic weights that connect the internal neurons poses one of the greatest (open) challenges. Since the spike trains of an SNN are formally represented as summations of Dirac delta functions, differentiation no longer applies, quickly ruling out the use of backprop.\footnote{Though there is no lack of effort in attempting to manipulate backprop to work with SNNs \cite{bohte2002error,lee2016training}. We also remark that other efforts often attempt to convert backprop-trained ANNs to non-trained equivalent SNN models \cite{diehl2016conversion}.}
One common, biologically-plausible rule used to adjust SNN synapses is spike-timing dependent plasticity (STDP) \cite{bi1998synaptic}, which takes into account the temporal ordering of spikes in order to conduct either long-term potentiation (LTP), where a weight's strength (or its efficacy) is increased, or long-term depression (LTD), where a weight's strength is decreased. However, STDP rules are typically applied over a temporal window, similar in spirit to backprop through time (BPTT) \cite{werbos1990backpropagation}, making it less useful for incremental, real-time learning, something that the human brain is clearly able to do. Although STDP can be formulated to operate incrementally (without a window) through the use of traces, online adjustment of SNN weights still remains an extremely difficult challenge, and, in the case of SNNs with more than one layer, STDP and related approaches often struggle to incrementally train the model properly. 

In order to tackle the difficult problem of effectively and efficiently learning multiple layers of spiking neurons, we will start by first reformulating the SNN architecture under the \emph{spiking neural coding} framework, which will bridge the recently proposed discrepancy reduction family of learning algorithms \cite{ororbia2017learning} with core computational principles of networks of spiking neurons \cite{maass1997networks,eliasmith2012large}. Discrepancy reduction algorithms fundamentally embody the neuro-mechanistic theories of predictive coding \cite{friston2009predictive,bastos2012canonical,rao1999predictive}, prospective coding \cite{rainer1999prospective,komura2001retrospective}, and analysis by synthesis \cite{neisser2014cognitive}. The central idea behind predictive coding (or predictive processing \cite{clark2015surfing}) is that the brain could be viewed as a top-down, directed generative model that first actively generates hypotheses (or predictions) and then immediately corrects itself in the presence of environmental stimuli. In the statistical learning literature, this theoretical framework has been concretely implemented and extended to include how synaptic weight adjustment is conducted locally, recently branded as neural generative coding (NGC) \cite{ororbia2019lifelong,ororbia2022neural} \footnote{This naming also highlights the fact that this approach is not the same as the classical signal processing usage of the phrase ``predictive coding'' \cite{atal1979predictive}, which itself has also seen recent some integration into SNN learning and inference \cite{boerlin2013predictive}.}. One of the primary contributions of this work is to recast the core principle components and ideas behind SNNs into the framework of NGC. Unlike most current SNNs, which are designed to operate like typical feedforward or recurrent ANNs, we will show that the iterative nature of NGC is naturally suited to continuous-time presentation of input stimuli, allowing it to perform continual error-correction \cite{ororbia2020continual} in order to build up dynamic distributed representations of data from spike trains. Furthermore, by doing so, the NGC-reformulated SNN can easily exploit a simple, neurocognitively-plausible, local learning rule, 
which could be used in place of or in tandem with STDP.

\section{Methodology} 
\label{sec:continual_learning_spikes}

\subsection{Problem Definition: Continually Predicting Sensory Stream Inputs}
\label{sec:problem}
In this article, the neural agents we are ultimately concerned process sensory stimuli, which are sampled from an environment or a stochastic process over time. The patterns presented to the neural system are to be viewed as a continuum, i.e.,  $\{(\mathbf{y}_1,\mathbf{x}_1, \mathcal{T}_1) \dots (\mathbf{y}_{n},\mathbf{x}_{n}, \mathcal{T}_n)\}$ with $n$ examples, though in the case of most real-world application streams, there would be no finite bound on the number of samples. At any particular instant, $\mathbf{x}_j \in \mathcal{R}^{D \times 1}$ represents the $D$-dimensional feature vector of the $j^\text{th}$ example and $\mathbf{y}_j \in \mathcal{R}^{C \times 1}$ is the target -- in the case of classification, it is instead $\mathbf{y}_j \in \{0,1\}^{C \times 1}$ where $C$ is the number of classes). 
$\mathcal{T}_n$ is an optional task descriptor (which we will use later in our continual learning experiments). Note that the agent, in the stream setting, is to process any instance only once as it continues through the data continuum, meaning that no exact data point previously seen can be retrieved for additional repeated processing (unless it is stored in a small replay buffer) as is typically done when training most ANNs. The main task for any learner receiving samples from such a continuum is to extract, as quickly as possible, enough information about the (nonstationary) distribution that governs the observations received in order to effectively predict either the input stimulus $\mathbf{x}_j$ itself or its target $\mathbf{y}_j$. Other tasks are possible, e.g., the learner could ideally learn how to generate $\mathbf{x}_j$ over a finite time horizon which is characteristic of the generative modeling of a time-series and would be useful in supporting higher-level cognitive activities such as planning in robotics.

The data stream formulation presented above still treats each data point as a discrete element although SNNs are intended to process input stimuli that arrive within the flow of continuous time. In order to truly adapt the above stream to SNNs, we further generalize the setting such that each tuple $(\mathbf{y}_j, \mathbf{x}_j)$ is presented to the learner for a fixed period of simulated time. More importantly, the vectors in this tuple are mapped to an appropriate spike train encoding, e.g., a Poisson spike train. In essence, a data tuple is to be presented to the agent for a specified stimulus interval of $T_{st}$ milliseconds (ms), optionally followed by an inter-stimulus time of $T_{ist}$ ms (where no input is provided). The length of the stimulus and inter-stimulus times will vary depending on the data problem and application, e.g., frames of a video might be presented for a shorter stimulus interval ($30-50$ ms) while static images might be presented for a longer period ($100$-$200$ ms). Inter-stimulus times in this work will be $100-150$ ms in order to allow the neurons to decay back to their resting potentials. 

\subsection{Spiking Neural Coding: Continual Error-Correction from Spike Trains}
\label{sec:spike_ncn}
Next, we present a mechanistic description of our proposed spiking neural coding framework for handling the continuous-time data prediction problem described above.
In principle, the spiking neural coding network (SpNCN), the model instantiation of our framework, could be designed to work with any type of simulated neuron, including complex, more biologically-faithful ones, e.g., Hodgkin–Huxley \cite{hodgkin1952quantitative} or Izhikevich \cite{izhikevich2003simple} neurons. However, in this study, we will formulate the system using the leaky integrate-and-fire (LIF) neuron, one of the simpler yet faster-to-simulated spiking unit models. Though it ignores many biophysical details of the act of spiking, the LIF model is computationally cheaper to simulate and captures the core intuition behind how neurons behave in the presence of input \cite{oreilly2000computational,eliasmith2004neural}. In the LIF, the rate at which spikes are produced is positively correlated with the strength of the current that enters a cell.\footnote{In addition, note that we are furthermore simulating point approximations of neural cells, which exploit the fact that the electrical signals that propagate from the dendrites to the cell body are essentially averaged together \cite{oreilly2000computational}. } 

To describe the simulation of an SpNCN, we next develop the important model equations for inference and learning in matrix-vector form. We will start from a neuronal unit-agnostic form of the SpNCN -- meaning that any type of neuronal unit, including the Hodgkin-Huxley unit, could be integrated and used -- before specifying the exact specification of the LIF variant that we will experiment with in this work. As is the case with determining the exact specification of rate-coded NGC-based models used in machine learning literature \cite{ororbia2019lifelong,ororbia2022neural}, the general SpNCN can be defined in terms of three key computations: 
1) prediction or hypothesis generation, 
2) error-correction, and 
3) weight adaptation. 

\paragraph{Prediction and Error Correction:} From the perspective of spike-trains, the first two computations listed above can be described in terms of how they affect the input current $\mathbf{j}^\ell(t) \in \mathcal{R}^{D_\ell \times 1}$ that is fed into a block of $D_\ell$ cells (at layer $\ell)$. Consider the case of an SpNCN with $L$ layers of cells (one layer of sensory/actuary cells and $L-1$ layers of internal processors -- each layer contains $D_\ell$ units), where we specify the (binary) spike vectors for any layer as $\mathbf{s}^\ell(t)$ and their filtered transformations (or traces) as $\mathbf{z}^\ell(t)$. The spike vectors are produced by a spike-response model (SRM) $( \mathbf{v}^\ell(t) , \mathbf{s}^\ell(t) ) \leftarrow f_{srm}(\mathbf{v}^\ell(t), \mathbf{j}^\ell(t))$, which is a function that takes in an input current and the state of the voltage variable block that we are interested in tracking. The SRM, which we will leave unspecified for now, captures the essence of most neuronal behaviors (including refractoriness) one would like to model given limited computational resources available for simulation.
To compute the filtered transform of a binary spike vector $\mathbf{s}^\ell_t$, we make use of a variable trace (often used in online implementations of STDP) defined as follows:
\begin{align}
\mathbf{z}^\ell(t) = \mathbf{z}^\ell(t) + \frac{\partial \mathbf{z}^\ell(t)}{\partial t}, \; \mbox{where, } \frac{\partial \mathbf{z}^\ell(t)}{\partial t} = -\frac{\mathbf{z}^\ell(t)}{\tau_{tr}} + \mathbf{s}^\ell(t)
\end{align}
which, in effect, applies a low-pass filter to the spike trains in a layer $\ell$. Each dimension of the trace $\mathbf{z}^\ell(t)$ is increased by one at the moment of a spike recorded in $\mathbf{s}^\ell(t)$ and decreases exponentially with time constant $\tau_{tr}$. 
Importantly, using a low-pass/trace filter smooths out the sparse spike trains generated by the raw SRM while still being biologically-plausible, and this operates similarly to a rate-coded equivalent value that is maintained by the actual neuronal cells, possibly in the form of the concentration of internal calcium ions \cite{oreilly2000computational} (for example, a variable trace could correspond to the glutamate bound to synaptic receptors).

\begin{figure*}
\begin{center}
\includegraphics[width=\textwidth]{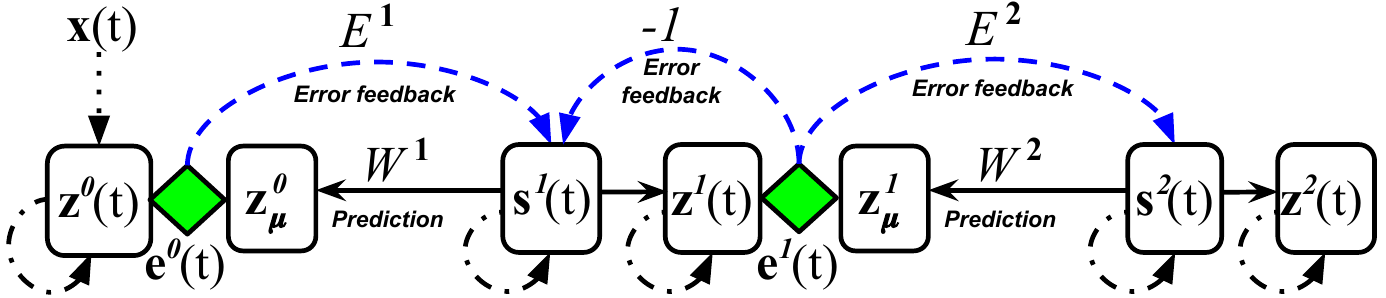}
\caption{A $2$-layer spiking neural coding network architecture. Green diamonds indicate error units ($\mathbf{e}^0(t), \mathbf{e}^1(t)$), which, in this simplified model depiction, compute the amount of mismatch between the current predictions ($\mathbf{z}^0_\mu, \mathbf{z}^1_\mu$) and the target signal traces ($\mathbf{z}^0(t), \mathbf{z}^1(t)$). Variables $\mathbf{s}^0(t), \mathbf{s}^1(t)$ represent the binary spike outputs 
of a particular vector grouping of neuronal units at time $t$. Black dash-dotted arrows represent recurrent transmission of the last known value of the relevant variable. Mismatch signals are carried back across sets of (a mixture of excitatory and inhibitory) synapses, depicted by blue dashed arcs, which are used to adjust the current action potential of the spiking neurons. Solid black lines depict predictive/generative synapses while black dotted lines just indicate direct passing of information. }
\label{fig:spncn}
\end{center}
\vspace{-0.5cm}
\end{figure*}

In order to compute the input (electrical) current values needed for an SRM, we must first specify the first two computations of the SpNCN. 
Starting from the spike variable blocks $\{ \mathbf{s}^1_t, \mathbf{s}^2_t \}$, which are output from the underlying SRMs, the SpNCN makes local predictions, in parallel, as follows:
\begin{align}
    \mathbf{z}^\ell_\mu = \mathbf{W}^\ell \cdot \mathbf{s}^\ell(t), \quad \mathbf{e}^\ell(t) = \Big( \mathbf{z}^\ell(t) - \mathbf{z}^\ell_\mu \Big) \label{eqn:error}
\end{align}
where the (predictive) synaptic weights for a layer $\ell$ are organized into a matrix $\mathbf{W}^\ell$. Note that $\cdot$ denotes a matrix/vector multiplication. Note that we model the output layer predictions and error units of the SpNCN a bit differently, i.e., we treat them as stateful spiking neurons (see Appendix for details). 
Mismatches between these predictions and currently-existing filtered state variables are then computed by an additional, coupled set of error neurons $\mathbf{e}^\ell(t)$, as depicted in the equation above. In the brain, our claim is that, much in the same vein as \cite{friston2009predictive}, error neurons (and the error weights) can be related to superficial pyramidal cells which pass along mismatches signals while deep pyramidal cells pass along predictions from state units.\footnote{Note that in \cite{friston2009predictive}, the message passing done by superficial pyramidal cells is referred to as ``backwards'' transmission while the deep pyramidal cell message passing is referred to as ``forwards'' transmission. However, in the SpNCN/NCN, these directions are ``flipped'' -- our error weights do the work of ``backwards'' transmission and the prediction weights do the work of ``forwards'' transmission. Nonetheless, for the SpNCN framework, it is more intuitive to instead think of message passing as moving vertically where prediction weights generate top-down expectations and error weights relate corrective signals, similar in spirit to \cite{rao1999predictive}.}
These error neurons, key to the inherent continual error-correction process of the SpNCN,  
are now to be treated in continuous-time as well.
When these error cells are connected recurrently to neurons in a nearby layer, which introduces two new additional sets of synaptic weights $\{ \mathbf{E}^1, \mathbf{E}^2 \}$ (the error synapses), we may compute the input current to each layer as in the following differential equations:
\begin{align}
    \tau_j \frac{\partial \mathbf{j}^\ell(t)}{\partial t} &= -\kappa_j \mathbf{j}^\ell(t) + \phi_e \big( -\mathbf{e}^\ell(t) + \mathbf{E}^\ell \cdot \mathbf{e}^{\ell-1}(t) \big)\\
    \tau_j \frac{\partial \mathbf{j}^L(t)}{\partial t} &= -\kappa_j \mathbf{j}^L(t) + \phi_e \big( \mathbf{E}^L \cdot \mathbf{e}^{L-1}(t) \big)
\end{align}
which yields, when integrating with the forward Euler method, the following updates:
\begin{align}
    \mathbf{j}^\ell(t + \Delta t) &= \mathbf{j}^\ell(t) + \frac{\Delta t}{\tau_j} \Big( -\kappa_j \mathbf{j}^\ell(t) + \phi_e \big( -\mathbf{e}^\ell(t) + \mathbf{E}^\ell \cdot \mathbf{e}^{\ell-1}(t) \big) \Big) \label{eqn:curr} \\
    \mathbf{j}^L(t + \Delta t) &= \mathbf{j}^L(t) + \frac{\Delta t}{\tau_j} \Big( -\kappa_j \mathbf{j}^L(t) + \phi_e \big( \mathbf{E}^L \cdot \mathbf{e}^{L-1}(t) \big) \Big) \label{eqn:curr_top}
\end{align}
where $\kappa_j$ is a coefficient to control the strength of the conductance leak and  $\phi_e(\cdot)$ is a nonlinear transform applied to the error message input pool (in this work, we use the identity $\phi_e(\mathbf{v}) = \mathbf{v}$).
Notice that, unlike in standard spiking neural networks, which are typically designed to be feedforward in nature, the SpNCN operates like a recurrent top-down generative model where error messages are propagated across time through the recurrent synapses.

The general key components of the framework specified above are illustrated in Figure \ref{fig:spncn} (and depicted in Algorithm 1 in the Appendix). All that remains is to define the third computation of an SpNCN, which, interestingly enough, can still be done without having yet specified a concrete SRM. 

\paragraph{Synaptic Updates through Spike-Triggered Local Representation Alignment: }
\label{sec:weights}
In the presence of time-varying input, we propose adapting the synaptic connections of the SpNCN using a new form of the coordinated local learning rule, local representation alignment (LRA) \cite{ororbia2017learning,ororbia2018conducting,ororbia2019biologically}, that has been successfully applied to modern-day, second generation artificial neural systems. This learning rule, which we will call \emph{Spike-Triggered LRA} (ST-LRA), is an event-driven update which means that the presence of a binary spike will trigger an adjustment of the synapses for the relevant predictor neurons. Given that the predictions within the neural system are made in parallel to one another, once a hypothesis/prediction for any particular layer has been made, the corresponding error neurons $\mathbf{e}^\ell_t$ are able to immediately perform the needed mismatch comparison (between the current state of the target area and prediction). The nature of the model's prediction mechanism also entails parallel computation of weight updates -- once the mismatch signal has been computed (Equation \ref{eqn:error}), a weight update for a layer $\ell$ of neurons may be readily computed. 

Much like a Hebbian update rule \cite{hebb1949organization}, ST-LRA makes use of readily available local information, making it biologically-plausible and much more hardware-friendly. The predictive weights for any layer $W^\ell$, as well as its corresponding set of error weights $E^\ell$, are updated as follows:
\begin{align}
    \Delta \mathbf{W}^\ell = \mathbf{e}^{\ell-1}(t) \cdot (\mathbf{s}^\ell(t) )^T, \quad \Delta \mathbf{E}^\ell = \beta \Big(\mathbf{s}^\ell(t) \cdot (\mathbf{e}^{\ell-1}(t) )^T \Big)
\end{align}
where $\beta$ is a coefficient that controls how quickly the error weights evolve, and usually can be set to a value close to one, such as $\beta = 0.9$. 
Interestingly enough, this LRA update
could be viewed to work similarly to the classical delta rule \cite{widrow1960adaptive} and the prescribed error rule \cite{macneil2011pes,bekolay2013simultaneous}. However, using an LRA update in a neural model means that one is committed to accepting that there are neurons that are tasked solely with mismatch computations. Fortunately, in the brain, evidence of the existence of these error neurons has been found, especially in visual cortical circuits \cite{miller1993activity,brown2001recognition,bell2016encoding,zmarz2016mismatch,fiser2016experience}.
Furthermore, much in line with the hard/soft weight bounding employed in existing spike-timing rule simulations, we control for potential weight value explosion by bounding the Euclidean-norm of the vector columns of each weight matrix to a maximum length of $20$.
Biologically, this type of normalization could exist as a result of limited resource-availability which drives the actual process of synaptic adjustment. 
Once updates are computed, weight matrices would be updated via a single weighted step (akin to stochastic gradient ascent): $\mathbf{W}^\ell \leftarrow \mathbf{W}^\ell + \alpha_u \Delta \mathbf{W}^\ell$ and $\mathbf{E}^\ell \leftarrow \mathbf{E}^\ell + \alpha_u \Delta \mathbf{E}^\ell$. $\alpha_u$ is the step size used in taking a step in the direction of the synaptic displacement. 

The synaptic evolution of the SpTNCN, under ST-LRA, is markedly different from spike-timing dependent plasticity (STDP) \cite{magee1997synaptically,bi1998synaptic,bi2001synaptic}, which is one carefully studied rule used in adjusting synaptic efficacies in SNNs. However, although the SpNCN uses a discrepancy reduction-based approach to adaptation, there is no reason why the ST-LRA update could not also be combined with the online form of STDP. Given that the brain is likely to employ multiple, different kinds of adjustment rules, doing so might allow for the fusion of both generative and discriminative knowledge when rapidly processing time-varying data points, which aligns with previous work that has empirically shown the benefits of doing so in rate-coded statistical learning setups \cite{oreilly1998sixprinciples,oreilly2000computational,ororbia_deep_hybrid_2015a,ororbia2015learning,ororbia2020continual,ororbia2019lifelong}.
For example, one can combine ST-LRA with online STDP through a convex combination 
to create the following two-term rule:
\begin{align}
    \Delta \mathbf{W}^\ell = (1 - \lambda) \Big( \mathbf{e}^{\ell-1}(t) \cdot (\mathbf{s}^\ell(t))^T \Big) - \lambda \Big( A_+(\mathbf{W}^\ell) (\mathbf{s}^{\ell-1}(t) \cdot (\mathbf{z}^\ell(t))^T) \nonumber\\
    + A_-(\mathbf{W}^\ell) (\mathbf{z}^{\ell-1}(t) \cdot (\mathbf{s}^\ell(t))^T) \Big)
\end{align}
where $\lambda$ is a coefficient introduced to control the strength of the STDP term throughout synaptic weight evolution. $A_+(W^\ell)$ and $A_-(W^\ell)$ are elementwise functions meant to introduce the soft (or hard) weight bounds generally used when training pure STDP-based SNNs. Furthermore, notice that the trace variable $\mathbf{z}^\ell$ is reused for the STDP co-term.

With hypothesis generation, error-correction, and weight adaptation fully specified (see Appendix for pseudocode), all that remains is to decide on an SRM $f_{srm}(\cdot)$ to completely define a concrete SpNCN model. In the appendix, we present a full algorithmic specification for a general SpNCN that uses any kind of SRM neuronal model.
We will next describe the SRM that we will focus on in this paper. However, once again, we note that other choices of more intricate, bio-physical SRMs might facilitate different behavior and more closely mimic other important properties of biological neural circuitry.

\begin{figure*}
\begin{center}
\includegraphics[width=\textwidth]{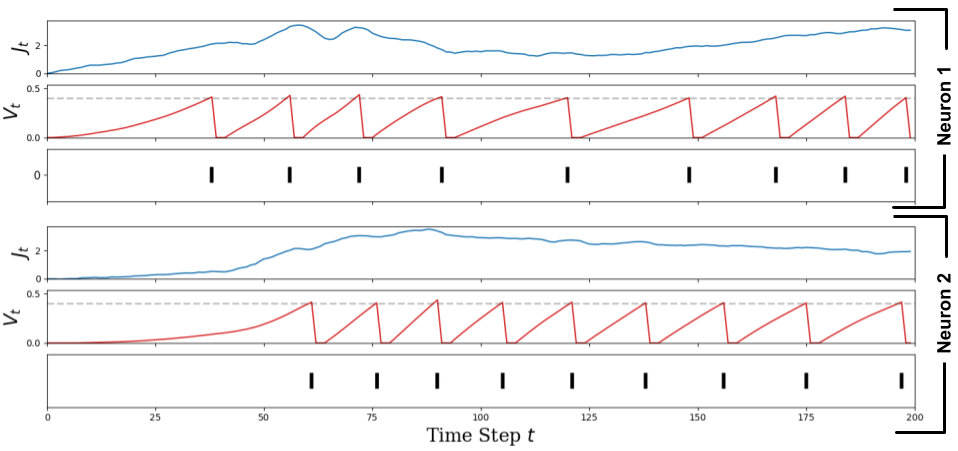}
\caption{A demonstration of a spike trains generated by two randomly sampled LIF neurons within an SpNCN trained on the MNIST dataset. The model in this example was presented with the digit ``8''. Plots include, from top to bottom (for each of the two neurons, i.e., ``Neuron 1'' and ``Neuron 2''), the synaptic conductance $\mathbf{j}_t$ over time (ms), the membrane potential, and the resulting raster (spike) plot (each solid vertical dash marks mean the neuron's membrane potential exceeded the threshold $\mathbf{v}_{thr}$ (depicted by a horizontal dashed light gray line).
}
\label{fig:spike_train}
\end{center}
\vspace{-0.5cm}
\end{figure*}

\paragraph{The Leaky Integrate-and-Fire SRM: } 
\label{sec:srm}
The SRM $f_{srm}(\mathbf{v}^\ell(t), \mathbf{j}^\ell(t))$ we implemented is the leaky integrate-and-fire (LIF) \cite{eliasmith2004neural} process, which entails modeling neurons as leaky integrators of their current inputs. This integration over time, for a block/layer $\ell$ of neurons, is specified by the following differential equation:
\begin{align}
    \tau_m \frac{\partial \mathbf{v}^\ell}{\partial t} = -\gamma_m \mathbf{v}^\ell(t) + R_m \mathbf{j}^\ell(t) \label{eqn:lif_diff}
\end{align}
where $R_m$ is the corresponding membrane resistance (for any single cell in the entire system), $\tau_m$ is the membrane time constant (specifically set as $\tau_m = R_m C_m$, where $C_m$ is the membrane capacitance), and $\gamma_m$ is a coefficient that controls the strength of the leak. The above equation, in actuality, describes a simple resistor-capacitor (RC) circuit, with the leakage resulting from the resistor. The current $\mathbf{J}^\ell(t)$ is integrated over time by the capacitor, which is placed in parallel to the resistor. In order to simulate leaky integrator neuron dynamics with Equation \ref{eqn:lif_diff}, we approximate the differential equation with the forward Euler method (as was done for the current in Equations \ref{eqn:curr}-\ref{eqn:curr_top}) to compute the value of the voltage $\mathbf{v}^\ell(t)$. This calculation proceeds as follows:
\begin{align}
    \mathbf{v}^\ell(t + \Delta t) = \mathbf{v}^\ell(t) + \frac{\Delta t}{\tau_m} \Big( -\gamma_m \mathbf{v}^\ell(t) + R_m \mathbf{j}^\ell(t) \Big)
\end{align}
where $\Delta t$ is the integration time constant (replacing the $\partial t$ in the original differential equation). Notice that this formulation of the LIF assumes a resting (membrane) potential of $0$ decivolts (although we note that the SpNCN could be formulated with other bounds and threshold values).

In order to generate a spike, a threshold $\mathbf{v}_{thr}$ must be chosen to compare $\mathbf{v}^\ell(t)$ against. As the voltage $\mathbf{v}^\ell(t)$ accumulates over time, a neuron will emit a spike once its voltage exceeds $\mathbf{v}_{thr}$ (in the simulations of this paper, we operate in the $[0,1]$ decivolt range). Upon emitting a spike (yielding a value of $1$), a neuron's voltage variable is reset to $0$, our chosen voltage resting rate (or resting membrane potential) for this paper's experiments. A spike vector is simply created via the following vector elementwise comparison: $\mathbf{s}^\ell(t) = \mathbf{v}^\ell(t) \geq \mathbf{v}_{thr}$. 
In Figure \ref{fig:spike_train}, we depict the spike generation process (as a function of the modeled synaptic conductance and the membrane potential) for two sampled LIF units, i.e., ``Neuron 1'' and ``Neuron 2'', in an SpNCN trained on the MNIST database.

Finally, our implementation of the LIF SRM incorporates an absolute refractory period. Specifically, after an LIF neuron spikes, a period of $T_{r}$ milliseconds (ms) worth of simulation time must pass before a build-up of potential is allowed again and, in the meantime, the value of the neuron is fixed to its resting potential ($0$ decivolts) until the refractory period has ended. In this work, we fix the absolute refractory period to be $T_{r} = 1$ ms (unless noted otherwise).


\paragraph{Input Representation: }
\label{sec:inputs}
To present data to the SpNCN, the input patterns that are clamped to its bottom-layer sensors are transformed on-the-fly to adhere to Poisson spike-trains. In the case of images, which could be randomly shuffled static pictures or ordered frames of a video, two-dimensional (2D) pixel grids are flattened to 1D vectors. The values of each pattern vector $\mathbf{x}$ are then normalized to lie in the range $[0,1]$ by dividing the scalar values along each dimension by the maximum pixel value $255$ (since pixels naturally live in the range $[0,255]$). To obtain the final Poisson spike rates, the normalized pattern vectors are then scaled such that the input firing rates are within the approximate range of $0$ to $63.75$ Hertz (Hz).
Then, given the scaled firing rates, we sample the input firing rates to obtain spike vectors at any step in time.
Note that, as it is done for the internal layers of the SpNCN, a trace filter is applied to the spike train stimulus.

\section{Experimental Results}
\label{sec:experiments}
In this section, we present two sets of experiments that test the SpNCN's ability to generate and classify patterns as well as its ability to retain knowledge over task sequences (see the Appendix for two additional preliminary experiments).

\subsection{Model Simulations}
\label{sec:simulations} 

\paragraph{Image Categorization Tasks:}
\label{sec:classification}
We start by investigating a high-dimensional classification challenge using the MNIST database\footnote{
http://yann.lecun.com/exdb/mnist/.}. This dataset contains images of $28\times28$ gray-scale pixel grids (feature values in the range of $[0,255]$).
Fashion MNIST \cite{xiao2017fashion}, on the other hand, contains $28x28$ grey-scale images of 10 classes of clothing items instead of digits or characters (data setup is the same as the one for MNIST). We also investigated the Stanford Optical Character Recognition (OCR) \footnote{
http://ai.stanford.edu/$\sim$btaskar/ocr/} and Caltech 101 Silhouettes \footnote{
https://people.cs.umass.edu/$\sim$marlin/data.shtml} datasets, both of which were in binary vector form and thus were not normalized (before converting to Poisson spike trains).
For all datasets, the image vectors were converted to Poisson spike trains via the process described in the last section such that the maximum spike rate was $63.75$ Hz (to be in accordance with \cite{diehl2015unsupervised}). 
Images were presented for a stimulus time of $T_{st} = 200$ ms.

\begin{table}
\caption{Generalization error (lower is better) of spiking networks on MNIST. The performance of our SpNCN and implemented SNN baselines (labeled ``impl.'') were measured over $10$ trials (mean and standard deviation are reported). Models below the double line are online models.}
\label{results:mnist}
\begin{center}
 \begin{tabular}{l | l c c} 
 \hline
 \textbf{Model}  & Preprocess? & Type & Performance \\ 
 \hline
 Dendritic Neurons \cite{hussain2014improved} & Yes & Rate-based & $9.7$\% \\ 
 Spiking RBM \cite{merolla2011digital} & No & Rate-based & $11.0$\% \\
 Spiking RBM \cite{oconnor2013real} & Yes & Rate-based & $5.9$\% \\
 Spiking CNN, BP \cite{diehl2015fast} & No & Rate-based & $0.9$\% \\
 Spiking RBM \cite{neftci2014event} & Yes & Rate-based & $7.4$\% \\
 Spiking RBM \cite{neftci2014event} & Yes & Spike-Based & $8.1$\% \\
 Spiking CNN \cite{zhao2014feedforward} & Yes & Spike-Based & $8.7$\% \\
 2-Layer SNN \cite{brader2007learning} & Yes & Spike-Based & $3.5$\% \\
 ML H-SNN \cite{beyeler2013categorization} & Yes & Spike-Based & $8.4$\% \\
 2-Layer SNN \cite{querlioz2013immunity} & No & Spike-Based & $6.5$\% \\
 2-Layer SNN \cite{diehl2015unsupervised} & No & Spike-Based & $5.0$\% \\
 syn-SNN (STDP) \cite{hao2018biologically} & No & Rate-based & $3.27$\% \\
 SNN-LM \cite{hazan2018unsupervised} & No & Spike-Based & $5.93$\% \\
 \hline
 \hline %
 2-Layer SNN, $3$ passes \cite{diehl2015unsupervised} & No & Spike-Based & $\sim 17.1$\% \\
 Online SNN-LM \cite{hazan2018unsupervised} & No & Spike-Based & $6.61$\% \\
 Online SCNN \cite{thiele2018event} & No & Spike-Based & $4.76$\% \\
 SNN, df-DRTP \cite{frenkel2019learning} (impl.) & No & Spike-Based & $40.13 \pm 0.35$\% \\
 SNN, df-BFA \cite{samadi2017deep} (impl.) & No & Spike-Based & $9.38 \pm 0.12$\% \\
 SpNCN (ours) & No & Spike-Based & $\mathbf{2.47 \pm 0.16}$\% \\ 
 \hline
\end{tabular}
\vspace{0.2cm}
\caption{Samples of the SpNCN's averaged input predictions. }
\label{fig:spike_samples}
\begin{tabular}{m{1.5cm}|m{7.1cm}}
\textbf{Model} & \multicolumn{1}{c}{MNIST Samples} \\
\hline
\hline
SpNCN & \includegraphics[width=7cm]{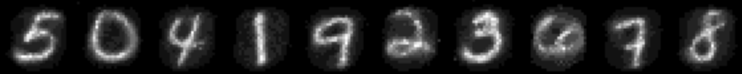} 
\end{tabular}

\caption{Generalization error (lower is better) of various online spiking networks on Fashion MNIST, Stanford OCR, and Caltech 101 datasets. The performance of the SpNCN and implemented SNN baselines (the SNN, df-DRTP \& SNN, df-BFA) were measured over $10$ trials (mean and standard deviation are reported). Models below the double line are online models.}
\label{results:fmnist}
 \begin{tabular}{l | l c c c c} 
 \hline
   &  &  & Fashion MNIST & Stanford OCR & Caltech 101 \\ 
 \textbf{Model} & Preprocess? & Type & Performance & Performance & Performance\\ 
 \hline
 ANN (BP) & No & Rate-based & $12.98$\% \cite{ororbia2019biologically}  & $37.01 \pm 1.19$\% & $44.77 \pm 1.22$\% \\
 syn-SNN (STDP) & No & Rate-based & $15.35$\% \cite{hao2018biologically} & -- & -- \\
 \hline
 \hline
 SNN, df-DRTP \cite{frenkel2019learning} (impl.) & No & Spike-Based & $45.27 \pm 0.26$\% & $82.19 \pm 0.12$\% & $77.75 \pm 0.24$\% \\
 SNN, df-BFA \cite{samadi2017deep} (impl.) & No & Spike-Based & $25.65 \pm 0.12$\% & $84.11 \pm 0.51$\% & $63.68 \pm  0.42$\% \\ 
 SpNCN (ours) & No & Spike-Based & $15.61 \pm 0.05$\% & $42.78 \pm 0.16$\% & $51.67 \pm 0.53$\% \\
 \hline 
\end{tabular} 
\vspace{-0.5cm}
\end{center}
\end{table}

The SpNCN models that were simulated consisted of $3$ layers of LIF units (each with $3000$ units). The absolute refractory period was $1$ ms, $\kappa_j = 0.25$, the integration time constant was $\Delta t = 0.25$ ms, and the time constants were $\tau_j = 10$ ms (for the LIF SRM, we set $\tau_m = 20$ ms and $\mathbf{v}_{thr} = 0.4$ decivolts). The trace time constant was set to $\tau_{tr} = 30$ ms. Step size was $\alpha_u = 0.055$ and $\beta = 1.0$. Given that the SpNCN is unsupervised, we use the training labels after the SpNCN has seen the entire dataset once by fitting a maximum entropy classifier to approximate rate-code equivalents of its spike-trains (see the later Equation \ref{eqn:rate_code_convert}).
We furthermore implemented a standard SNN trained online with broadcast feedback alignment (BFA), specifically a derivative-free variant \cite{samadi2017deep} (\emph{df-BFA}) to ensure that the SNN can employ the exact same LIF SRM that the SpNCN uses. This baseline is referred to as \emph{SNN, df-BFA} in all tables. In additionally, we implemented a derivative-free variant of another broadcast alignment-like algorithm known as direct random target projection (\emph{df-DRTP}) \cite{frenkel2019learning} as an additional alternative to training SNNs. In the appendix, we describe the details of our \emph{df-BFA} and \emph{df-DRTP} implementations.


In Table \ref{results:mnist}, we report the generalization performance (on each dataset's respective test set), averaged over $10$ trails (mean and standard deviation are reported).
Note that in our evaluation, the SpNCN's reported generalization is the result of letting it adapt to the MNIST training set for only a single time, since our goal is to evaluate the model's online streaming performance exclusively. The spiking models we compare its performance are ones that were trained over multiple passes (epochs) over the database. For an approximation of how a performant SNN trained with STDP performs in a similar setting, we add a line to the table with the value explained in \cite{diehl2015unsupervised} (the original paper reported only $3$ passes over MNIST, which means it has a slight advantage over the SpNCN in that it see the MNIST data at least two more times).  The performance of an STDP-driven SNN is significantly lower than the SNN trained for $15$ epochs on MNIST (the last model performance just above the double line in Table \ref{results:mnist}), which was a wide model containing $6400$ neurons (technically $6400$ excitatory neurons and $6400$ inhibitory neurons), thus much wider than the streaming SpNCNs we experimented with in this study. We furthermore compare to powerful spiking convolutional networks (SCNNs) \cite{thiele2018event}. We remark  that the best performing model, i.e., the ``Spiking CNN, BP'', was trained with backprop and serves as an idealized upper bound given that the rest of the models (including the SpNCN) are adapted to data using biologically-plausible rules.

Observe that in the case of MNIST, our proposed SpNCN actually performs comparably to the most powerful (non-backprop-based) SNN models. This is interesting for two reasons: 1) the SpNCN's synaptic adjustments rely heavily on the simple ST-LRA update rule (using STDP as a small regularizer if $\lambda > 0$) and 2) the SpNCN is trained online, unlike most of the models we compare to (which were trained by passing over the MNIST dataset multiple times). 
We should note that some state-of-the-art performance for SNNs on MNIST include $2.8$\% \cite{mozafari2019bio} and $1.6$\% \cite{kheradpisheh2018stdp}. However, both of these models (and many like them) are convolutional in nature and make use of additional complex learning mechanisms such as reinforcement learning. In terms of fully-connected SNN architectures, the SpNCN is quite competitive (even as an online learning algorithm) and, notably, outperforms the  more complex SCNN \cite{thiele2018event}. Note that it would be fruitful to incorporate additional learning rules from recent work \cite{kheradpisheh2018stdp,mozafari2019bio} to improve the SpNCN's generalization ability further. As shown earlier, ST-LRA could be integrated with other learning rules/complementary mechanisms such as STDP.
In Figure \ref{fig:spike_samples}, we visualize SpNCN predictions of randomly sampled digits (images show spike predictions averaged across the stimulus time).  



Finally, in Table \ref{results:fmnist} we report training the same 3-layer SpNCN on Fashion MNIST, Stanford OCR, and Caltech 101. While current research has not focused as much on evaluating SNNs on Fashion MNIST as much (and has not examined Stanford OCR and Caltech 101 to our knowledge), we provide, for reference, a backprop-trained rate-coded ANN. For Fashion MNIST, we note that although the SpNCN does not exactly match the non-spiking model's performance (on any dataset), it does come close and is competitive with the syn-SNN (which is impressive for a more difficult benchmark like Fashion MNIST).
Furthermore, our results show that the SpNCN consistently outperforms other SNNs trained with \emph{df-BFA} and \emph{df-DRTP} on the Stanford OCR and Caltech 101 datasets. 

\paragraph{On Catastrophic Forgetting:}
\label{sec:forgetting}
When trained on more than one dataset, or ``task'', sequentially, ANNs are known to succumb to catastrophic interference (or forgetting), where recently-seen information blindly overwrites prior knowledge already encoded in a neural system's synapses \cite{mccloskey_catastrophic_1989,ratcliff_connectionist_1990,lewandowsky1994relation}. Although a great deal of work exists in attempting to combat this difficult problem, however, the ANNs investigated rely on continuously graded and often heavily dense activity patterns when processing each and every pattern.
The resulting ``cross-talk'' between an ANN's neurons is often depicted to be one source of catastrophic interference and, classically \cite{french1999catastrophic}, as well as recently \cite{ororbia2019lifelong}, sparsity has been argued to be a key biologically-inspired solution to improve ANN memory retention. 
Since spiking neural architectures naturally make use of very sparse codes, we are interested in determining if this sparsity can be exploited to mitigate the amount of forgetting that a neural system experiences when adapting online to a sequence of different tasks. 

To investigate how the SpNCN handles data organized in groups/tasks, we examine the SpNCN under three conditions: 1) the basic SpNCN model (from the last experiment), 2) the SpNCN integrated with a small memory buffer (with $1000$ stored tuples) that replays previously encountered samples from prior tasks over time (we refer to this model as SpNCN-Buf), and 3) the SpNCN with an extra input that encodes the current task descriptor/identifier to create a task context that induces a form of lateral inhibition that makes it harder for non-task relevant neurons to fire (i.e., the lateral inhibition makes it harder for the membrane potential of non-task units to reach their spike thresholds) allowing the SpNCN system to decide on a number of units to allocate per task (we refer to this model as SpNCN-Lat). The first two models have the advantage that they do not require any additional information in the form of task identifiers while the third model does not require storing input patterns into a memory module. Given that it is likely that raw/natural sparsity alone will not effectively reduce neural cross-talk alone (unless perhaps if the system simulated were designed with a massive number of neural units), the last two cases allow us to explore how the SpNCN's sparsity might mix favorably with simple control structures that could be approximated by other neural circuits, e.g., the buffer samples in the SpNCN-Buf could come from long-term episodic memory while the task contexts in the SpNCN-Lat could come from signals (that selectively modulate activity in other brain areas) produced by either the basal ganglia, with evidence demonstrating its role as an information router \cite{yehene2008basal,buschman2014goal}, or the prefrontal cortex, which has been shown to induce a form of cognitive control \cite{miller1993activity,herd2014neural}. Furthermore, each SpNCN in this experiment contains an additional output neuron branch for predicting the target labels (see Appendix for details).

\begin{table*}[!t]
\caption{Memory retention ability measured on the SpNCN on Split MNIST, Split Fashion MNIST (FMNIST), and Split NotMNIST benchmarks. Mean (for ACC \& BWT) and standard deviation (for ACC) reported over $10$ trials.}
\label{results:lifelong_learning}
\begin{center}
\begin{tabular}{ c||c|c|c|c|c|c } 
 \hline
  &\multicolumn{2}{c|}{\textbf{MNIST}}&\multicolumn{2}{c}{\textbf{NotMNIST}}&\multicolumn{2}{c}{\textbf{FMNIST}}\\
 \hline
 \textbf{Model} & \textbf{ACC} & \textbf{BWT} & \textbf{ACC} & \textbf{BWT} & \textbf{ACC} & \textbf{BWT}\\  
 \hline
 EWC \cite{kirkpatrick2017overcoming} & $0.190 \pm 0.030$ & $-0.357$ & $0.186 \pm 0.020$ & $-0.361$ & $0.199 \pm 0.06$ & $-0.350$ \\
 SI \cite{zenke17synaptic_continual} & $0.197 \pm 0.110$ & $-0.367$  & $0.161 \pm 0.030$ & $-0.370$ & $0.198 \pm 0.100$ & $-0.370$ \\
 Lwf \cite{lwf} & $0.846 \pm 0.340$ & $-0.120$ & $0.626 \pm 0.091$ & $-0.130$ & $0.875 \pm 0.300$ & $-0.130$ \\
 IMM \cite{forgetting_iim} & $0.951 \pm 0.018$ & $-0.007$ & $0.925 \pm 0.011$ & $-0.006$ & $0.950 \pm 0.013$ & $-0.005$\\
 GDumb \cite{prabhu2020gdumb} & $0.978 \pm 0.09$ & $\mathbf{-0.005}$ & $0.940 \pm 0.080$ & $-0.004$ & $0.973 \pm 0.09$ & $-0.006$ \\
 \hline
 \hline
 SpNCN & $0.735 \pm 0.154$ & $-0.302$ & $0.776 \pm 0.228$ & $-0.228$ & $0.8324 \pm 0.097$ & $-0.198$ \\
 SpNCN-Buf & $0.943 \pm 0.451$ & $-0.020$ & $0.927 \pm 0.331$ & $-0.008$ & $0.951 \pm 0.329$ & $-0.028$  \\
 SpNCN-Lat & $0.972 \pm 0.297$ & $-0.001$ & $0.948 \pm 0.311$ & $-0.003$ & $0.985 \pm 0.216$ & $-0.001$  \\
 \hline
\end{tabular}
\end{center}
\vspace{-0.5cm}
\end{table*}

\begin{figure*}[!t]
\begin{center}
\caption{t-SNE visualizations of an SpNCN with access to the MNIST training set up front (left), a continual SpNCN-Buf (middle), and a continual SpNCN-Lat (right) (for MNIST). }
\label{fig:tsne}
\includegraphics[width=0.3225\textwidth]{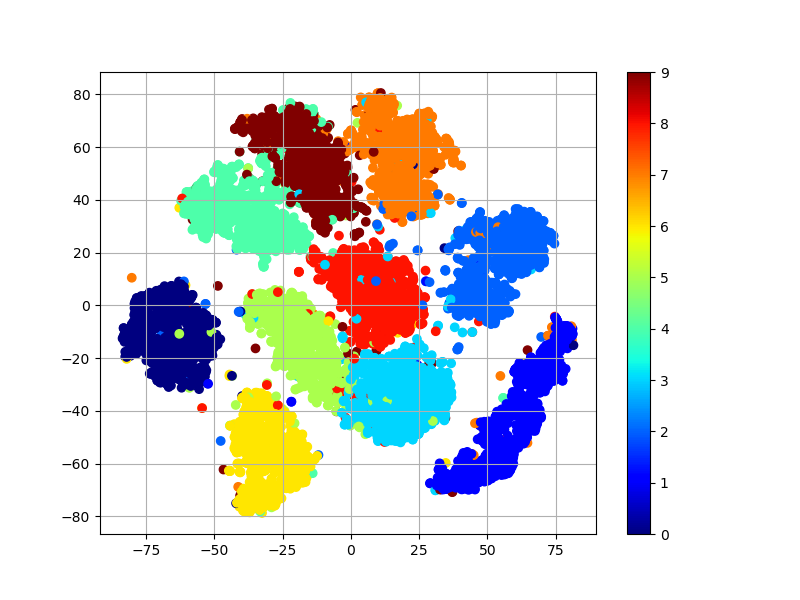}
~
\includegraphics[width=0.3225\textwidth]{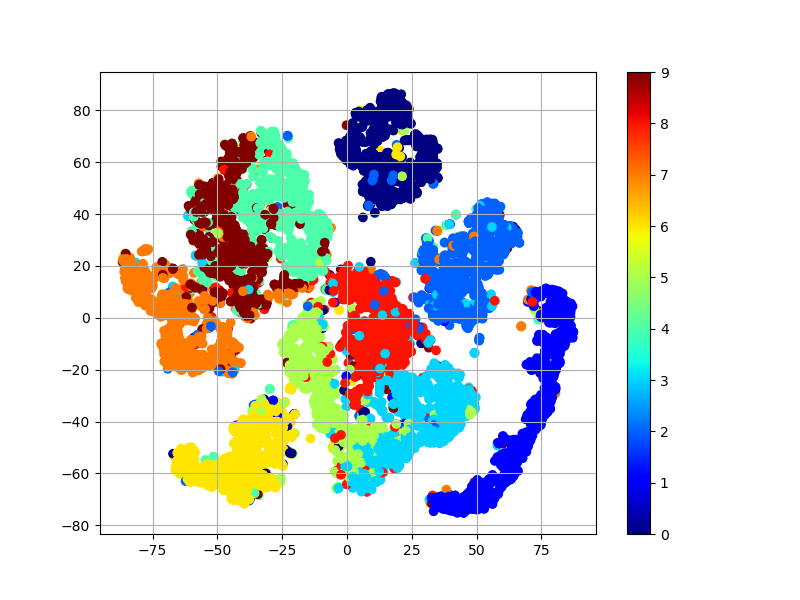}
~
\includegraphics[width=0.3225\textwidth]{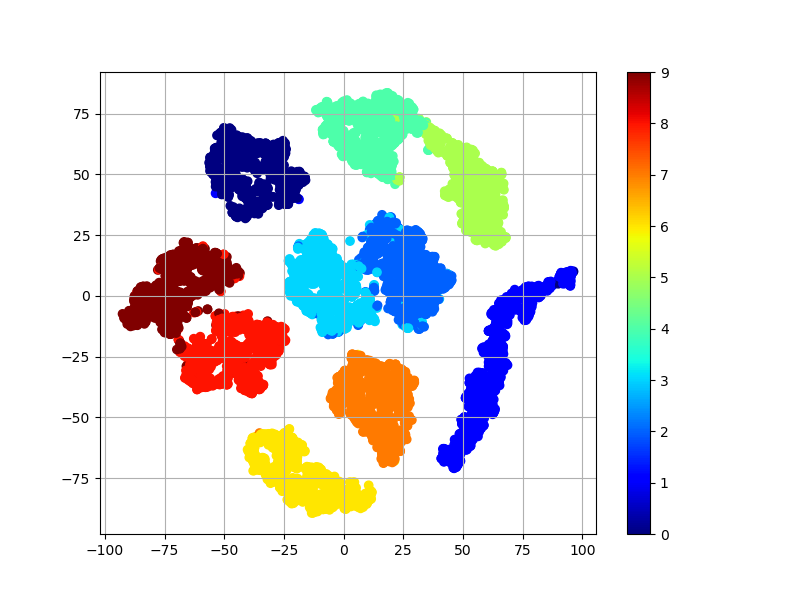}
\end{center}
\vspace{-0.5cm}
\end{figure*}

To experimentally evaluate the SpNCN's online continual learning ability, 
we adapt the Split MNIST,  NotMNIST, and Split Fashion MNIST (FMNIST) continual learning benchmarks to the spike-train continuous time setting. Split MNIST is effectively just the MNIST database (described earlier) but with the sample images grouped into five different subsets of consecutive digit pairs, where each subset (or task) corresponds to learning to distinguish between two different objects. 
The subsets are arranged in a particular order, i.e., a task sequence stream, presenting a particularly difficult learning challenge to the learner since there is cross-talk between digit classes in the output layer due to changes in the label distribution over subsets of digit samples. Split NotMNIST and Split FMNIST are identical to Split MNIST except that, instead of digits, the data subsets are either characters of varying fonts/glyphs (letters A-J) in the case of NotMNIST\footnote{
http://yaroslavvb.blogspot.com/2011/09/notmnist-dataset.html}, or clothing items in the case of FMNIST.

We evaluate performance based on average accuracy (ACC), which measures a model's overall generalization across a sequence of tasks after reaching the end of the data continuum stream, and backwards transfer (BWT), which measures the influence that learning a particular task has on those that were encountered before it (a higher BWT is better and a very negative BWT indicates a greater degree of forgetting). 
We report our ACC and BWT measurements in Table \ref{results:lifelong_learning} for all benchmarks.
We add as a reference the performance of several continual learning benchmark models/methods for rate-coded ANNs including elastic weight consolidation (EWC) \cite{kirkpatrick2017overcoming}, synaptic intelligence (SI) \cite{zenke2017continual}, Lwf \cite{lwf}, the mean incremental moment matching method (IMM) (Mean-IMM) \cite{forgetting_iim}, and the greedy sampler and dumb learner (GDumb) \cite{prabhu2020gdumb}.
We emphasize that our goal is not to necessarily reach state-of-the-art ANN ACC levels on these benchmarks but, rather, to experimentally determine if a spiking system can be designed to reduce cross-task forgetting.

As observed in Table \ref{results:lifelong_learning}, we see that neural processing based on the SpNCN's spike trains does indeed appear to positively reduce forgetting, though not as strongly as predicted. However, when this sparsity is integrated with simple structures such as a simple memory module or task context modulated lateral inhibition, we see that memory retention across tasks is significantly improved, resulting in performance that is competitive with modern-day rate-coded continual learning, backprop-based systems. With respect to both ACC and BWT, we see that the SpNCN-Lat experiences minimal forgetting when trained, resulting in performance that is competitive with GDumb \cite{prabhu2020gdumb}. However, in further contrast to the powerful baselines we compare to, \textbf{the SpNCN obtains competitive cross-task performance even though it is trained online}, i.e., it only processes data patterns once whereas, to obtain top performance, the rate-coded baselines were trained on each task for multiple epochs.

Finally, in order to qualitatively examine the quality of the spike-train representations acquired by SpNCNs trained in the continual learning scenario versus having access to the full dataset (as in the last experiment), we visualize the approximate layerwise embeddings of the SpNCN (from the first set of experiments) as well as the SpNCN-Buf and SpNCN-Lat (from the second set of experiments) when they are presented with (MNIST) test images. Specifically, we extract a vector embedding (of the layer farthest from the sensory input layer) for an image by computing the rate-code equivalent of the corresponding spike-train, for any particular layer $\ell$, via the following:
\begin{align}
    \mathbf{c}^\ell = \frac{\gamma_c}{N_c} \sum^{T_{st}}_{t=1} \mathbf{s}^\ell(t), \mbox{where, $0 < \gamma_c \le 1$} \label{eqn:rate_code_convert}
\end{align}
where $\gamma_c$ is a coefficient used to control how well the approximate encoding fits into the range of $[0,1]$ (we set $\gamma_c = 1$ for this study) and $N_c$ is the number of cycles over $T_{st}$, or the total number of steps taken within the stimulus time window. The resulting rate-code equivalent embeddings are plotted in Figure \ref{fig:tsne} using t-SNE to perform the mapping from the layer's original dimension to 2D (t-SNE was ran with a perplexity target of $30$ for $100$ iterations after an initial PCA projection). 
We observe that, while the SpNCN trained on the full MNIST dataset acquires representations that cluster together nicely, the SpNCN-Lat learns clusters that are more strongly separated (largely owing to the fact that it learns context information specific to each task). The SpNCN-Buf also learns clustered representations, however, these appear to messier and exhibit overlapping, which corroborates our quantitative finding that it exhibits more forgetting than the SpNCN-Lat.

\subsection{On Limitations}
\label{sec:discuss}
Although the proposed SpNCN represents an important step towards the bridging of practical machine learning and computational neuroscience, there are many aspects of our model, in its current form, that differ a bit from known neurobiology. While these elements can be addressed, incorporating them might introduce further challenges in developing a model that generalizes well while still adhering to the constraints that real biological neurons obey/satisfy.
One such element is the fact that excitatory and inhibitory synaptic weights in our SpNCN are not distinctly separated and are essentially ``globbed'' together. In the brain, it is a common pattern for there to only be a certain proportion of neurons that are inhibitory (about $20$\%) while the rest are excitatory. Furthermore, we remark that, while learning with ST-LRA update approach, the signs of the SpNCN's weights could very possibly change or flip during the course of model evolution. 
To remedy this, the SpNCN's input current variable block $\mathbf{j}$ could instead be formulated to separate out the excitatory and inhibitory input channels. In addition, a constraint could be imposed that fixes the signs of synapses at model initialization with a biologically realistic proportion of weights assigned negative signs. Further inspiration from neurocognitive architectures such as LEABRA \cite{oreilly2000computational} or spiking neural simulators such as BRIAN \cite{goodman2009brian} could help in cleanly integrating these design elements into the SpNCN computational framework.

While the LIF SRM we experimented with is quite useful and fast to simulate, especially with the inclusion of the absolute refractory period, there are still additional mechanisms that should be modeled in order to build a proper neuronal simulation. Beyond the inclusion of a relative refractory period, modeling other important neuronal dynamics such as accommodation (a neuron fatigues, or becomes less and less active for the same excitatory input) and hysteresis (a neuron remains active for a period of time even if the excitatory input fades or is removed) would prove interesting and could be done by modifying the LIF differential equation we specified earlier \cite{oreilly2000computational}. Alternatively, one could  swap out the LIF SRM used within the SpNCN with a more realistic (though computationally more expensive to simulate) neuronal model  \cite{izhikevich2003simple,hodgkin1952quantitative}. 

\section{Conclusion}
\label{sec:conclusion}
In this paper, we proposed spiking neural coding, a model that is composed of leaky integrate-and-fire neurons, and its learning algorithm for adjusting synaptic weight strengths incrementally. The model was formulated under a general framework that can  accommodate simulated neuronal models of varying complexity, ranging from the leaky integrate-and-fire unit (used in this study) to those based on the Hodgkin-Huxley coupled differential equations. Pattern classification and continual learning experiments demonstrate the generalization ability of our model in the realm of online, continuous-time based learning.

The spiking neural coding network works by continuously chasing targets, constantly making guesses of the activities of its internal neuronal units and then correcting its states to ensure that better predictions are made in the future.
Even though it is trained purely online, the model is competitive with many powerful SNNs learned with STDP or other mechanisms, including those with various incarnations of backpropagation of errors. Furthermore, our model exhibits better computational economy and is more biologically-plausible than modern-day (non-spiking) artificial neural networks.
Future work will entail investigating the further development of the proposed spiking network for use in more complex applications, especially those involving motor control, which would be of immense value to robotics research. In addition, the direct integration and evaluation of the proposed spiking system on neuromorphic hardware, where the energy efficiency gains will be most prominent and critical, will be explored.

\section*{Acknowledgements}
We would like to thank Chris Eliasmith, Terrence Stewart, and Peter Blouw for useful discussions on the internal mechanics of NENGO and the dynamics of spiking neurons, which helped in shaping and inspiring many of the ideas behind the custom code/implementation that drives this paper. We also thank Alexander Ororbia (Sr) for valuable comments/suggestions for the early drafts of this paper.

\bibliographystyle{acm}
\bibliography{ref}

\newpage
\section*{Appendix}
\label{sec:appendix}

\subsection*{Spiking Neural Coding Algorithmic Specification}
\label{sec:algo}
In this section, we present the full pseudocode, i.e., Algorithm \ref{algo:spncn_process}, for computing state updates and synaptic weight updates in an $L$-layered spiking neural coding network (SpNCN) at any point in continuous time.
Note that additional coefficients/hyper-parameters are input to the SpNCN's routines, i.e., $\{ \kappa_j, \alpha_j, \Delta t, \tau_j \}$ to control the electrical current modeling, $\{ \mathbf{v}_{thr}, \tau_m \}$ to control the SRM, $\tau_{tr}$ to control the trace variables, and $\{ \beta, \alpha_u \}$ to control the synaptic updates (with the optional $\lambda$ if STDP is to be integrated into the synaptic updates), and are values set by the user at the start of simulation (the meaning of each is described in the main paper).
The pseudocode presented below demonstrates the operation of a general SpNCN that could use any kind of spike response model (SRM), so long as the SRM can make use of the variable $\mathbf{J}^\ell(t)$ (current) in the form developed in this paper. Possible choices for the SRM could be variations of the leaky integrate-and-fire (LIF) model, the Hodgkin–Huxley \cite{hodgkin1952quantitative} model, or the Izhikevich \cite{izhikevich2003simple} model.

\begin{algorithm*}
	\caption{Process for SpNCN inference/synaptic evolution (at time $t$).}
	\label{algo:spncn_process}
	\begin{algorithmic}[1]
	    \State {\bf Inputs:} $\mathbf{x}(t)$, $\Theta = \{W_1,E_1,\cdots,W_L,E_L\}$, $\Upsilon = \{\mathbf{j}^1(t),\cdots,\mathbf{j}^L(t) \}$, 
	    \State \hspace{0.5cm} $\mathcal{E} = \{\mathbf{e}^1(t),\cdots,\mathbf{e}^L(t) \}$, $\mathcal{V} = \{\mathbf{v}^1(t),\cdots,\mathbf{v}^L(t) \}$, $\mathcal{S} = \{\mathbf{s}^1(t),\cdots,\mathbf{s}^L(t) \}$, 
	    \State \hspace{0.5cm} $\mathcal{Z} = \{ \mathbf{z}^1,\cdots,\mathbf{z}^L \}$
	    \State {\bf Hyperparameters:} $\{ \kappa_j, \alpha_j, \Delta t, \tau_j \}$, $\tau_{tr}$, $\{ \beta, \alpha_u \}$, $\{ \mathbf{v}_{thr}, \tau_m \}$ (for the SRM)
	    \LineComment Routine for computing variable state updates
	    \Function{ComputeStates}{$\mathbf{x}(t)$, $\mathbf{y}$, $\Theta$, $\Upsilon$, $\mathcal{E}$, $\mathcal{V}$, $\mathcal{Z}$ } 
	        \State $\mathbf{z}^0(t) = \mathbf{x}(t)$ \Comment{$\mathbf{x}(t)$ is a sample from an external Poisson spike train}
	        \State $\mathbf{j}^0_{\mu,x}(t + \Delta t) = \mathbf{j}^0_{\mu,x}(t) + \frac{\Delta t}{\tau_j} \Big( -\kappa_j \mathbf{j}^0_{\mu,x}(t) + \big( \mathbf{W}^1 \cdot \mathbf{s}^1(t) \big) \Big)$
	        \State $(\mathbf{v}^0_{\mu,x}(t), \mathbf{s}^0_{\mu,x}(t)) = f_{srm} \big( \mathbf{v}^0_{\mu,x}(t), \mathbf{j}^0_{\mu,x}(t) \big)$
	        \For{$\ell = 1$ to $L$ }
	            \State $\mathbf{j}^\ell(t) \leftarrow \Upsilon[\ell]$, $\mathbf{v}^\ell(t) \leftarrow \mathcal{V}[\ell]$, $\mathbf{z}^\ell(t) \leftarrow \mathcal{Z}[\ell]$ \Comment{ Extract  variable states }
	            \State $\mathbf{e}^\ell(t) \leftarrow \mathcal{E}[\ell]$, $\mathbf{e}^{\ell-1}(t) \leftarrow \mathcal{E}[\ell-1]$ \Comment{ Extract error unit values }
	            \If{$\ell < L$} \Comment{ Update values of electrical currents}
    	            \State $\mathbf{j}^\ell(t + \Delta t) = \mathbf{j}^\ell(t) + \frac{\Delta t}{\tau_j} \Big( -\kappa_j \mathbf{j}^\ell(t) + \phi_e \big( -\mathbf{e}^\ell(t) + \mathbf{E}^\ell \cdot \mathbf{e}^{\ell-1}(t) \big) \Big) $
    	        \Else
    	            \State $\mathbf{j}^L(t + \Delta t) = \mathbf{j}^L(t) + \frac{\Delta t}{\tau_j} \Big( -\kappa_j \mathbf{j}^L(t) + \phi_e \big( \mathbf{E}^L \cdot \mathbf{e}^{L-1}(t) \big) \Big)$
	            \EndIf
	            \LineComment{ Compute voltage \& spike given an SRM, apply filter to spike}
	            \State $( \mathbf{v}^\ell(t) , \mathbf{s}^\ell(t) ) \leftarrow f_{srm}(\mathbf{v}^\ell(t),  \mathbf{j}^\ell(t))$, 
	            \quad
	            $\mathbf{z}^\ell(t) = \mathbf{z}^\ell(t) + \big( -\frac{\mathbf{z}^\ell(t)}{\tau_{tr}} + \mathbf{s}^\ell(t) \big)$
	        \EndFor
	        \LineComment{ Make top-down predictions \& compute error neurons}
	        \For{$\ell = L$ to $2$ }
	            \State $\mathbf{z}^{\ell-1}_\mu = \mathbf{W}^\ell \cdot \mathbf{s}^\ell(t), \quad \mathbf{e}^{\ell-1}(t) = \Big( \mathbf{z}^{\ell-1}(t) - \mathbf{z}^{\ell-1}_\mu \Big)$ 
	        \EndFor
	        \State $\mathbf{z}^0_\mu = \mathbf{s}^0_{\mu,x}(t), \; \mathbf{e}^0 = \Big( \mathbf{z}^0(t) - \mathbf{z}^0_\mu \Big)$
    	    \State \Return $\Upsilon$, $\mathcal{E}$, $\mathcal{V}$, $\mathcal{S}$ \Comment Return overwritten/updated variables
        \EndFunction
        
        \LineComment Routine for computing synaptic adjustments
        \Function{UpdateSynapses}{$\mathcal{S}, \mathcal{E}$, $\beta$} 
            \For{$\ell = 1$ to $L$ }
                \State $\mathbf{s}^\ell(t) \leftarrow \mathcal{S}[\ell]$, $\mathbf{e}^{\ell-1}(t) \leftarrow \mathcal{E}[\ell-1]$  \Comment Extract layer-wise statistics
                \State $\Delta \mathbf{W}^\ell = \mathbf{e}^{\ell-1}(t) \cdot (\mathbf{s}^\ell(t) )^T$,  \quad $\Delta \mathbf{E}^\ell = \beta \Big(\mathbf{s}^\ell(t) \cdot (\mathbf{e}^{\ell-1}(t) )^T \Big)$
                \State $\mathbf{W}^\ell \leftarrow \mathbf{W}^\ell + \alpha_u \Delta \mathbf{W}^\ell$, \quad $\mathbf{E}^\ell \leftarrow \mathbf{E}^\ell + \alpha_u \Delta \mathbf{E}^\ell$ \Comment{Gradient ascent}
            \EndFor
        \EndFunction
	\end{algorithmic} 
\end{algorithm*} 

For the prediction layers of the SpNCN, there are two designs that will highlight: 1) the unsupervised SpNCN for the benchmark experiments, and 2) the continual SpNCN model used for the lifelong learning experiments. For case 1, the output prediction layer of the SpNCN (the prediction of the sensory inputs), we ran the following dynamics:
\begin{align}
    \mathbf{j}^0_{\mu,x}(t + \Delta t) &= \mathbf{j}^0_{\mu,x}(t) + \frac{\Delta t}{\tau_j} \Big( -\kappa_j \mathbf{j}^0_{\mu,x}(t) + \big( \mathbf{W}^1 \cdot \mathbf{s}^1(t) \big) \Big),  \label{eqn:unsup_x_current} \\
    (\mathbf{v}^0_{\mu,x}(t), \mathbf{s}^0_{\mu,x}(t)) &= f_{srm} \big( \mathbf{v}^0_{\mu,x}(t), \mathbf{j}^0_{\mu,x}(t) \big) \\
    \mathbf{z}^0_\mu &= \mathbf{s}^0_{\mu,x}(t), \; \mathbf{e}^0 = \Big( \mathbf{z}^0(t) - \mathbf{z}^0_\mu \Big) \label{eqn:unsup_output}
\end{align}
where we see that the model's expectation of the sensory input is simulated as LIF neurons. 
For case 2, we modeled the prediction/expectation of the sensory input and the target labels, we employed the following dynamics equations:
\begin{align}
    \mathbf{j}^0_{\mu,x}(t + \Delta t) &= \mathbf{j}^0_{\mu,x}(t) + \frac{\Delta t}{\tau_j} \Big( -\kappa_j \mathbf{j}^0_{\mu,x}(t) + \big( \mathbf{W}^1_x \cdot \mathbf{s}^1(t) \big) \Big) \\ (\mathbf{v}^0_{\mu,x}(t), \mathbf{s}^0_{\mu,x}(t)) &= f_{srm} \big( \mathbf{v}^0_{\mu,x}(t), \mathbf{j}^0_{\mu,x}(t) \big) \label{eqn:sup_x_current} \\
    \mathbf{j}^0_{\mu,y}(t + \Delta t) &= \mathbf{j}^0_{\mu,y}(t) + \frac{\Delta t}{\tau_j} \Big( -\kappa_j \mathbf{j}^0_{\mu,y}(t) + \big( \mathbf{W}^1_y \cdot \mathbf{s}^1(t) \big) \Big) \\ (\mathbf{v}^0_{\mu,y}(t), \mathbf{s}^0_{\mu,y}(t)) &= f_{srm} \big( \mathbf{v}^0_{\mu,y}(t), \mathbf{j}^0_{\mu,y}(t) \big) \label{eqn:sup_y_current} \\
    \mathbf{z}^0_\mu &= \mathbf{s}^0_{\mu,x}(t), \; \mathbf{e}^0 = \Big( \mathbf{z}^0(t) - \mathbf{z}^0_\mu \Big), \; \mathbf{e}^0_y = \Big( \mathbf{y} - \mathbf{s}^0_{\mu,y} \Big)  \label{eqn:sup_output}
\end{align}
where we notice that we do not couple a set of error feedback synapses to transmit the label-specific error messages $\mathbf{e}^0_y$ back to the first latent layer, i.e., $\mathbf{e}^0_y$ is only used to compute the local update to the label prediction synapses $\mathbf{W}^1_y$.
Note that for the error neurons/units could be modeled in yet a more biologically-realistic fashion where each layer of mistmach activities could be modeled by two separate (signed) populations where one transmits via excitatory synapses and the other via inhibitory synapses. This would circumvent the requirement that error units must take on negative values (whereas the spiking neurons are strictly positive) -- we will explore this modeling modification in future work.




\subsection*{The X-O Pattern Classification Task}
\label{sec:exp1}
As a sanity check, we originally started by evaluating our proposed SpNCN on a simple pattern recognition task often used in evaluating the basic discriminative ability of SNNs in general \cite{christophe2015pattern}. The task consists of simply distinguishing between a circle (``O'') shape from an X-cross (``X'') binary pattern. The image patterns are $16\times16$ (binary) pixel grids, with backgrounds that are black (pixel value of $0$) and foreground elements that are white (pixel value of $255$).

Training the model consisted of presenting a sequence of X and O patterns for a fixed duration of time, i.e., $T_{st} = 60$ ms and $T_{ist} = 30$ ms (which is simulated by clamping the error neurons at the sensory/input layer to zero vectors within this inter-stimulus interval, allowing the internal neurons to reach their resting potential).
After a brief training period of presenting a few patterns, the SpNCN is evaluated on the following test sequence: $\{O,O,X,X,O,X,X,X,X,O,O,X,O\}$, where each pattern also presented for a stimulus time of $T_{st} = 100$ ms.
After presenting only a few of these toy patterns to the network, the SpNCN was able to reach $100$\% mean accuracy on the test sequence (averaged over $10$ trials). 

\subsection*{Continual Signal Chasing: The Bouncing Ball Problem Revisited}
\label{sec:exp3}
When processing sensory input vectors from a stream, the SpNCN is engaged in the process of ``signal chasing'', constantly predicting the sensory input it is about to receive and then error-correcting its internal state before receiving the next input. In most applications where sensory streams are prevelant, e.g., autonomous vehicles and robotics, this means the SpNCN will make relevant predictions 
based on a continual, shifting representation of its environment which is further encoded in binary spike train patterns. 

To evaluate how well the SpNCN chases signals in a data continuum, we adapt the classic bouncing ball problem often used to evaluate generative models in statistical learning research, such as temporal variants of Boltzmann machines \cite{taylor2007modeling,sutskever2009recurrent} and neural predictive coding models \cite{ororbia2017learning}. 
The problem entails predicting frames of pre-generated video snippets of the simulated rudimentary physics of three balls bouncing around in a box.
Specifically, we implement it as continuously running stochastic process to generate streams of user-specified length, the output of which might serve as a new simple benchmark for evaluating the online adaptivity of spiking neural models.
We set up our generator to create a single long-running video of $n=3$ balls bouncing around in a pixel grid forcing the neural system to learn how to predict/generate incoming frames given its current state. Each gray-scale video frame is $16x16$ pixels and each frame is converted to a $64$ Hz Poisson spike train.
For this experiment, we generated a small finite stream of $K = 2000$ frames. A $2$-layer SpNCN is adapted online using ST-LRA on the stream for the first $1000$ frames. For the last $1000$ frames, we deactivate the learning rule in order to test if the SpNCN is still able to generate patterns without synaptic weight adjustment. The simulation time step used for this experiment was $\Delta t = 0.1$ ms (simulation consisted of $600000$ 
\begin{wrapfigure}{r}{2.6in}
\centering
\vspace{-0.75cm}
\includegraphics[width=2.5in]{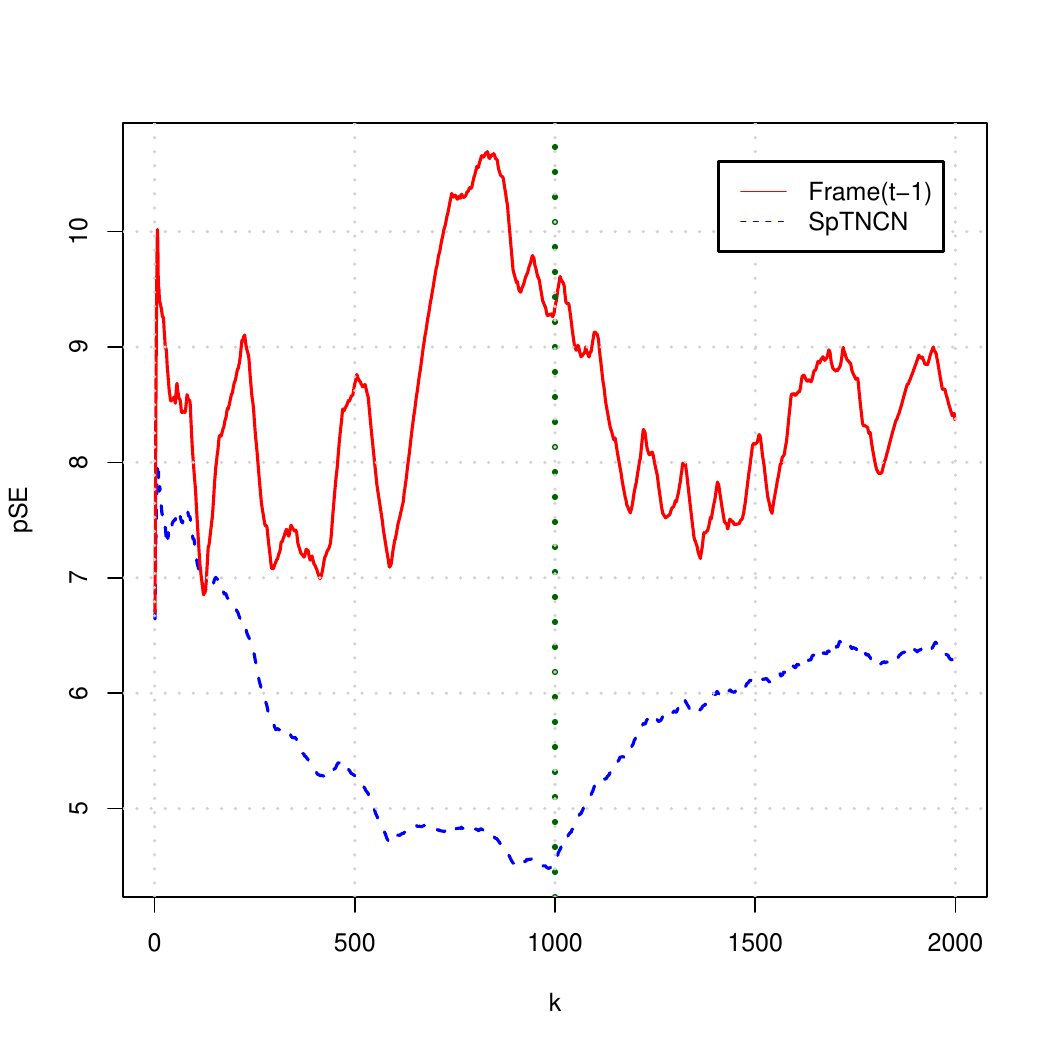}
\vspace{-0.4cm}
\caption{SpNCN tracking performance on the bouncing ball stream problem (pSE over video frame $k$). The vertical (green) dotted line marks the point when synaptic change was inhibited. }
\label{fig:signal_tracking}
\vspace{-0.45cm}
\end{wrapfigure}
discrete steps) and stimulus (video frame) presentation time was $T_{st} = 30$ ms. 
The SpNCN consists of $400$ LIF units in the first layer and $100$ units in the second layer. We use a leak scale of $\gamma = 0.25$ (and no refractory period for this simulation).

In online learning \cite{gama2013evaluating} it is common practice to record an evolving model's prediction error on a single sample presented at a particular instant in time and update a running, often decaying average of cumulative error. This is often referred to as prequential error \cite{gama2013evaluating,ororbia2015learning}. Our variant of this metric, prequential squared error (pSE), is calculated as follows:
\begin{equation}
\label{preq_error}
	 P_{\alpha}(i) = \frac{\sum^i_{k=1} \alpha^{i-k} \Big( (\mathbf{\widehat{x}} - \mathbf{x})^T \cdot (\mathbf{\widehat{x}} - \mathbf{x}) \Big)}{\sum^i_{k=1} \alpha^{i-k}}
\end{equation}
where $0 \ll \alpha \leq 1$ (we set $\alpha = 0.995$ for this experiment). $\mathbf{\widehat{x}}$ is the expected value of the SpNCN's prediction, $\EX[\mathbf{z}^0_{\mu}]$, of the $k$-th target video frame, estimated as the empirical average $\sum^{T_{st}}_{t=1} \mathbf{z}^0_{\mu}(t)$ over the stimulus presentation period.

In Figure \ref{fig:signal_tracking}, we plot the SpNCN's learning curve (pSE) across the entire simulated $30$ second period. To check that the SpNCN is indeed learning how predict its input over time, we also plot the pSE of the \emph{Frame(t-1)} model as a reference, which simply predicts the next frame by copying the exact previously seen frame (and often makes for a strong baseline model).
As observed in the figure, the SpNCN is able to maintain a reasonably good reconstruction of the target signal while consistently doing better than the baseline. The SpNCN reached an average $pSE = 6.672$ whereas the baseline predictor reaches a $pSE = 10.225$. More importantly, we should emphasize that over the $300000$ discrete simulation steps taken (for the $1000$ training frames, each presented $30$ ms), only $46,393$ synaptic weight updates were made for $\{W1,E1\}$ and only $28,998$ updates were made for $\{W2,E2\}$. This means that the number of weight adjustments is quite sparse over time which might prove to be a significant economy in the employment of real-time hardware systems. Furthermore, the fact that upper layer (second LIF layer) updates its synapses fewer times than the lower layer (first LIF layer) could be likened to the process of slow feature analysis \cite{wiskott2002slow}, since higher-level feature detectors would be (roughly) operating on a slower-moving time-scale than the lower-level units.

\subsection*{Derivative-Free Broadcast Feedback Alignment (df-BFA)}
\label{appendix:bfa}
In this paper, we train a spiking neural network (SNN) with broadcast feedback alignment (BFA), much as was done in \cite{samadi2017deep}. For direct comparability to our proposed spike-triggered local representation alignment for training systems of spiking neurons without derivatives, we specifically implemented and compared to a derivative-free variant of BFA (df-BFA). Focusing on this algorithm allows us to work with spike response models (SRMs) of any kind directly (such as the leaky integrate-and-fire, LIF, model of this paper and others). Furthermore, this algorithm does not require continuous approximations of activities, much as was needed in \cite{samadi2017deep}. This makes SNNs trained with BFA comparable to the spiking neural coding networks (SpNCNs) of this paper, and furthermore, more general.

Specifically, our implementation of \emph{df-BFA} was as follows, which we will describe using the same notation developed in this paper (which also makes it consistent with the machine learning research related to the general neural coding network \cite{ororbia2017learning,ororbia2020continual,ororbia2019lifelong}). The input current to any internal layer $\ell$ of a feedforward SNN trained by \emph{df-BFA}, is expressed as:
\begin{align}
    \mathbf{J}^\ell(t) = (1 - \kappa)\mathbf{J}^\ell(t) + \kappa \Big( -\gamma \mathbf{J}^\ell(t) + \phi( W^\ell \cdot \mathbf{s}^{\ell-1}(t) ) \Big)
\end{align}
which is then provided to the desired SRM, $(\mathbf{v}^{\ell}(t),\mathbf{s}^\ell(t)) \leftarrow f_{srm}(\mathbf{v}^{\ell}(t), \mathbf{J}^\ell(t))$. $\gamma$ controls the strength of the conductance leak, $\phi(\cdot)$ is a nonlinearity applied to the incoming signal pool (in this work, $\phi(v) = v$), and $\kappa$ is the conductance time constant. In this paper, we use the same LIF SRM used within the SpNCN.
To generate the necessary learning signals under \emph{df-BFA}, we introduce fixed, randomly initialized feedback (alignment) weights $F^\ell$ that directly connect the error neurons found at the output (top layer) of the network to each layer within the SNN. 
The output error signals are modeled by the equation $\mathbf{e}^L(t) = (\mathbf{\widehat{y}}(t) - \mathbf{y}) $ noting that, much like in \cite{samadi2017deep}, these error neurons could be made to be more biologically-realistic by creating two separate populations where one transmits via excitatory synapses and the other via inhibitory synapses. The update for the weights of layer $\ell$ is then:
\begin{align}
    \Delta W^\ell = \mathbf{d}^\ell \cdot (\mathbf{s}^{\ell-1})^T,  \mbox{where, } \mathbf{d}^\ell = \phi_e( F^{\ell} \cdot \mathbf{e}^L(t) ) \mbox{.}
\end{align}
Note that the top-layer/prediction weights would do not use the above equation, since they connect directly to the output units, and use an update calculated as: $\Delta W^L = \mathbf{e}^L \cdot (\mathbf{s}^{L-1})^T$. We extend the original BFA approach by incorporating what we call an ``error function'', $\phi_e(\cdot)$, which is an element-wise nonlinearity applied to the teaching signal vector(s). We found in our experiments that using this function improved the stability of BFA. We set $\phi_e(v) = sign(v)$, or the signum function.

Surprisingly, the \emph{df-BFA} scheme works quite well and, in our preliminary experiments, appeared to be reasonably robust to the initialization of the feedback weights $F^\ell$ (which are held fixed during learning). As a result, this simple synaptic update rule allows us to train SNNs of the same depth as the SpNCNs trained in this paper, yielding a strong baseline.

\subsection*{Derivative-Free Direct Random Target Propagation (df-DRTP)}
\label{appendix:drtp}
As another interesting baseline, we further implement the very recently proposed direct random target propagation algorithm and extend it by omitting its dependence on the activation function's first derivative so that way we may directly apply it to the same SNN architecture as we do for df-BFA. The underlying neural architecture is the same SNN as described in the section for BFA.

Under df-DTRP, which introduces fixed random projection synaptic matrices $P^\ell$, the updates for the weights of layer $\ell$ are as simply as follows:
\begin{align}
    \Delta W^\ell = \mathbf{d}^\ell \cdot (\mathbf{s}^{\ell-1})^T,  \mbox{where, } \mathbf{d}^\ell = \phi_e( P^{\ell} \cdot \mathbf{y} )
\end{align}
where we see that now the updates to the layerwise synapses are local in nature given that we now randomly project (with fixed initial values) the actual target label vector $\mathbf{y}$ to create a proxy training signal for any particular layer $\ell$. Note that $\phi_e(v) = sign(v)$, just as was done for BFA described earlier.

\end{document}